\documentclass[12pt]{article}
\usepackage{hyperref}
\usepackage{indentfirst}
\usepackage[english]{babel}
\usepackage[utf8]{inputenc}
\usepackage{csquotes}
\usepackage{array}          % Для настройки столбцов
\usepackage{booktabs}

\usepackage[letterpaper,top=2cm,bottom=2cm,left=3cm,right=3cm,marginparwidth=1.75cm]{geometry}

\usepackage{amsmath}
\usepackage{graphicx}

\title{\textbf{Track Anything Annotate: Video annotation and dataset generation of computer vision models.}}
\author{
    \textbf{Nikita Ivanov, Mark Klimov,} \\
    \textbf{Dmitry Glukhikh, Tatiana Chernysheva\footnote{For correspondence, contact (email: t.y.chernysheva@utmn.ru).}, Igor Glukhikh} \\
    \ University of Tyumen \\
}
\date{}

\begin{document}
\maketitle

\begin{abstract}
Modern machine learning methods require significant amounts of labelled data, making the preparation process time-consuming and resource-intensive. In this paper, we propose to consider the process of prototyping a tool for annotating and generating training datasets based on video tracking and segmentation. We examine different approaches to solving this problem, from technology selection through to final implementation. The developed prototype significantly accelerates dataset generation compared to manual annotation. All resources are available at \href{https://github.com/lnikioffic/track-anything-annotate}{https://github.com/lnikioffic/track-anything-annotate}.
\end{abstract}

\section{Introduction}

The development of neural networks and computer vision requires huge amounts of labeled data. However, their preparation remains a challenging task. Traditional manual annotation methods (e.g., using tools like LabelImg \cite{ke2024integrations} or VGG Image Annotator \cite{dutta2016vgg}) are time consuming and costly. This limits the scaling and deployment of state-of-the-art models in real-world applications. Manual partitioning may take too long in the data preparation phase to train models \cite{barskaya2024creation}. In addition, even semi-automated solutions relying on classical libraries like OpenCV \cite{sharma2021object} face accuracy problems in complex scenarios (e.g., when objects intersect or lighting conditions change).

To solve this problem, we decided to explore several tools for working with images and videos. The first tool was the OpenCV \cite{chandan2018real} library of computer vision algorithms with the CSRT tracker \cite{farhodov2019faster, farkhodov2020object}: Discriminative Correlation Filter (with Channel and Spatial Reliability). The Fast Segment Anything Model (FastSAM) \cite{zhao2023fast} began to be used to improve the detection accuracy. It is a CNN-based \cite{he2017mask} solution for the real-time Segment Anything task. Then, XMem \cite{cheng2022xmem} and XMem++ \cite{bekuzarov2023xmem++} were used instead of OpenCV to improve the quality of object tracking. It is a video object segmentation architecture for long videos with unified feature memory stores. FastSAM has been replaced by a segmentation model, Segment Anything Model 2 (SAM 2) \cite{ravi2024sam2}. It is a model for solving the segmentation problem in images and videos. SAM2 maintains hints and computes masks in real time for interactive use. This model is trained on the SA-V Dataset \cite{ravi2024sam2}.

The purpose of this paper is to prototype a 'Track Anything Annotate' tool to segment objects in video and then generate a data set for computer vision models that utilizes SAM2 and XMem++. Its creation will speed up the process of creating datasets.

The practical novelty lies in the combination of advanced models for segmentation and tracking. The prototype tool presented is unparalleled and directly addresses the problem of high labor intensity of annotation and scaling limitations of computer vision systems, offering an efficient alternative to traditional partitioning methods.

Thus, the prototype presented solves several problems at once:
\begin{enumerate}
    \item Reduce labor costs by automating repetitive operations.
    \item Improved accuracy through interactive correction and adaptation to complex scenarios (e.g. overlapping objects).
\end{enumerate}

This approach responds directly to the challenges described in project management studies: balancing quality, timing and budget. Unlike traditional methods, where a trade-off between speed and accuracy is inevitable, our tool offers a solution that combines the advantages of both approaches.

To achieve this goal, several automatic annotation methods were tested. The following combinations of technologies have been considered: OpenCV + FastSAM, FastSAM + XMem, SAM2 + XMem++, which will be discussed below.

\section{Methodology}

\subsection{OpenCV + FastSAM}

\textbf{OpenCV and CSRT Tracker \cite{farhodov2019faster}}. OpenCV is a library of computer vision algorithms for image processing. It has a CSRT tracker — discriminative correlation filter. However, this approach has several limitations: 1. Low accuracy when objects overlap or change position rapidly; 2. Tracking a large number of objects is difficult because a new tracker needs to be created for each new object. Computations are performed using the CPU, which cannot perform them in parallel.

\textbf{Fast Segment Anything Model \cite{zhao2023fast}}. CNN-based solution \cite{he2017mask} for the real-time Segment Anything task. FastSAM is designed to segment any object in an image. Significantly reduces computational requirements while maintaining competitive performance. This makes FastSAM a practical choice for a wide range of vision tasks.

\begin{figure}[htbp]
\centering
\includegraphics[width=0.8\linewidth]{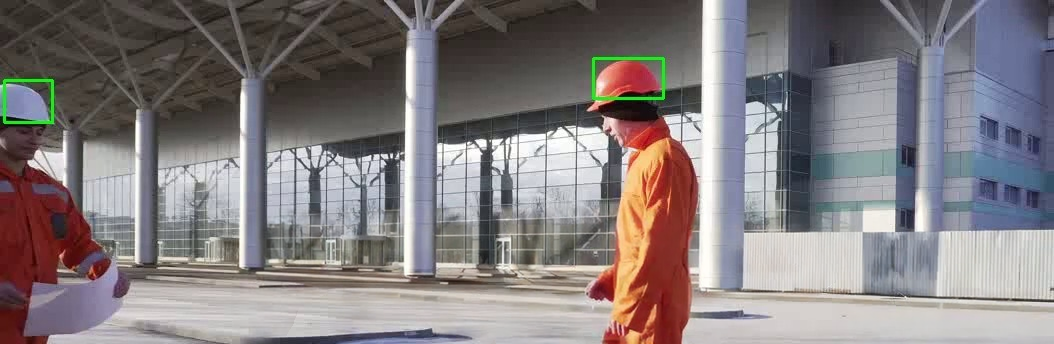}
\caption{\label{fig:1}Object selection.}
\end{figure}

\begin{figure}[htbp]
\centering
\includegraphics[width=0.8\linewidth]{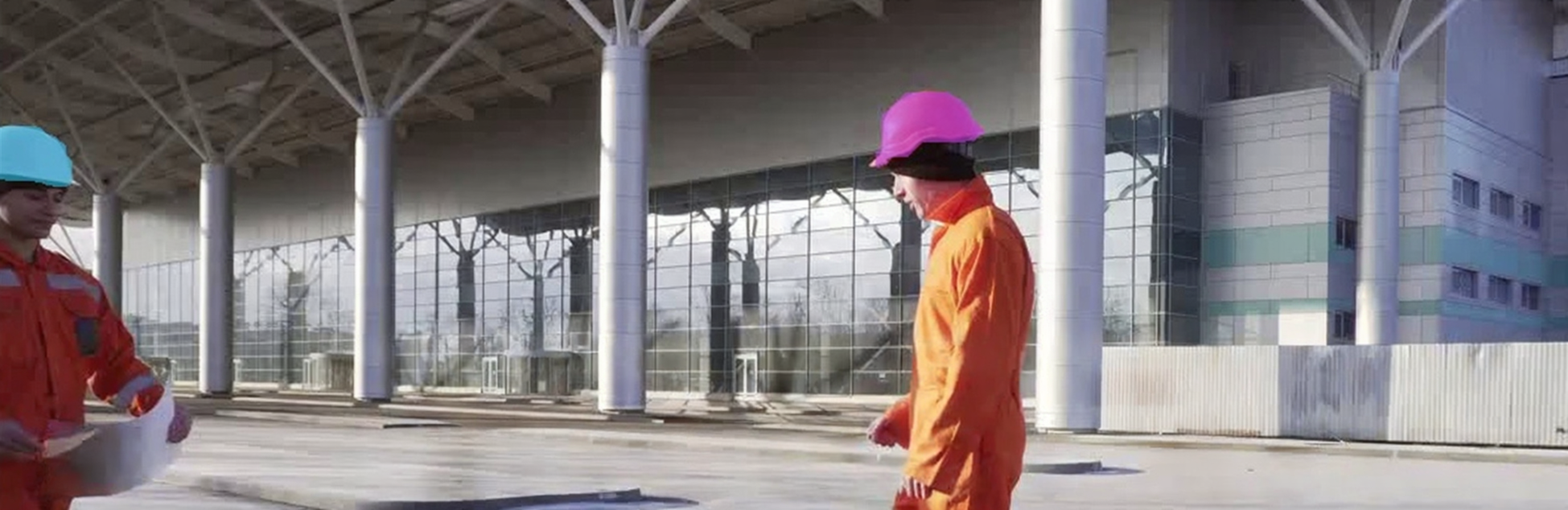}
\caption{\label{fig:2}Object segmentation.}
\end{figure}

\textbf{Object selection on video and segmentation.} Interactive object selection was limited by the accepted coordinate format of the CSRT tracker. The tracker only accepted the region of interest in the bounding-box format. A large number of objects could not be tracked. Figure \ref{fig:1} shows the object selection stage. Next, all the received frames with object coordinates are passed to FastSAM to segment the objects and obtain the object coordinates. The segmentation results are shown in Figure \ref{fig:2}.

\subsection{FastSAM + XMem}

This step was aimed at improving object tracking by replacing OpenCV with XMem. This step allowed for more accurate and flexible object tracking in the video. 

\begin{figure}[htbp]
\centering
\includegraphics[width=1\linewidth]{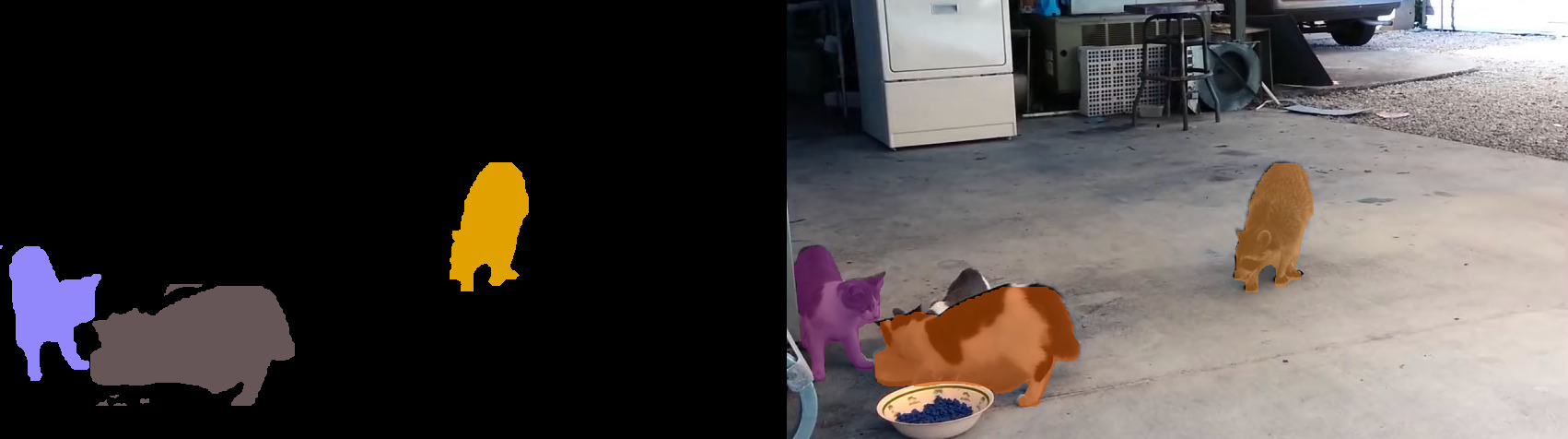}
\caption{\label{fig:3}Segmentation of the image with the resulting mask and the image with the mask applied.}
\end{figure}

\textbf{FastSAM \cite{zhao2023fast}} was used to segment the first frame of the video. The resulting mask was then passed to \textbf{XMem \cite{cheng2022xmem}}, which performed object tracking and generated masks on subsequent frames. Image segmentation can be performed using cues in the form of points or rectangles, allowing flexibility in defining the region of interest. Figure \ref{fig:3} shows image segmentation with the resulting mask.

\textbf{Image Segmentation.} Images can be segmented using dots or boxes. This gives you the flexibility to select objects in the first frame for later transfer to XMem.

\subsection{SAM2 + XMem++}

\textbf{Segment Anything Model 2 \cite{ravi2024sam2}.} An advanced tool from Meta AI. Designed for complex segmentation of objects in images and videos. It does an excellent job with complex visual data. All because of its unified model architecture with hinting capability. It supports real-time processing, zero-shot generalization, flexible use of hints, and computes masks in real-time. This allows SAM2 to be used interactively.

\textbf{XMem++ \cite{bekuzarov2023xmem++}.} Video segmentation tool. Accepts user-provided masks. XMem++ is built on top of Xmem. Improvements: 1. Persistent memory module that greatly improves model accuracy; 2. An algorithm for selecting candidates for annotation.

\begin{figure}[htbp]
\centering
\includegraphics[width=1\linewidth]{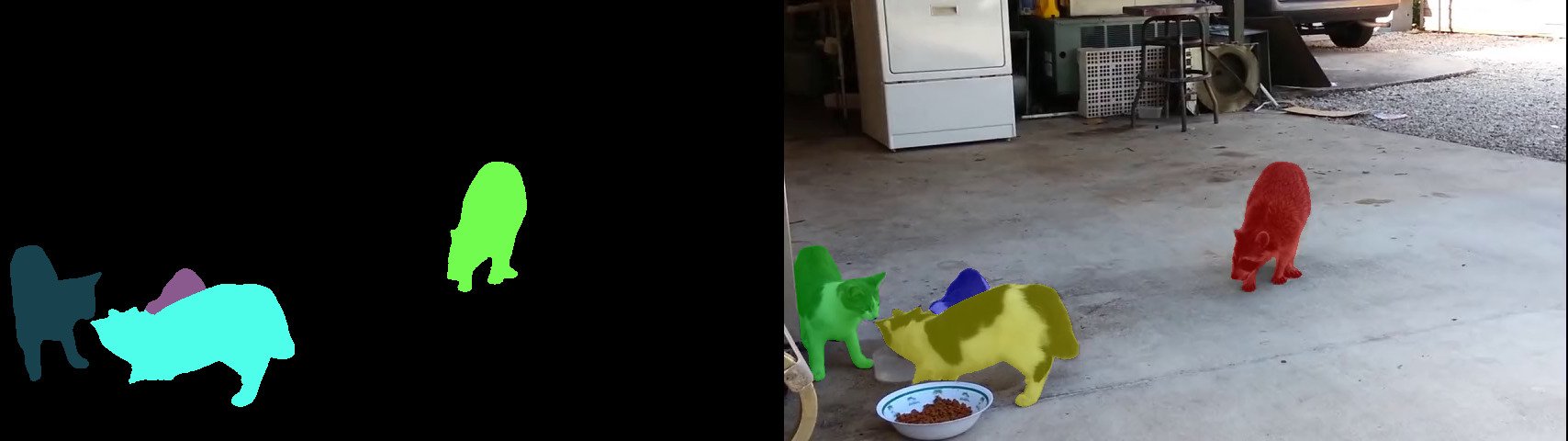}
\caption{\label{fig:4}Segmentation of the image with the received and applied mask when using SAM2.}
\end{figure}

\textbf{Advantages over the previous combination.} The accuracy and quality of segmentation have improved by using SAM2. This can be seen in Figure \ref{fig:4}. Shows the segmentation of the image with the received mask and the image with the superimposed mask. Compared to the segmentation with FastSAM, the contours of the objects are smoother and the amount of noise is reduced.

\subsection{Comparison of segmentation methods}
\begin{figure}[htbp]
\centering
\includegraphics[width=1\linewidth]{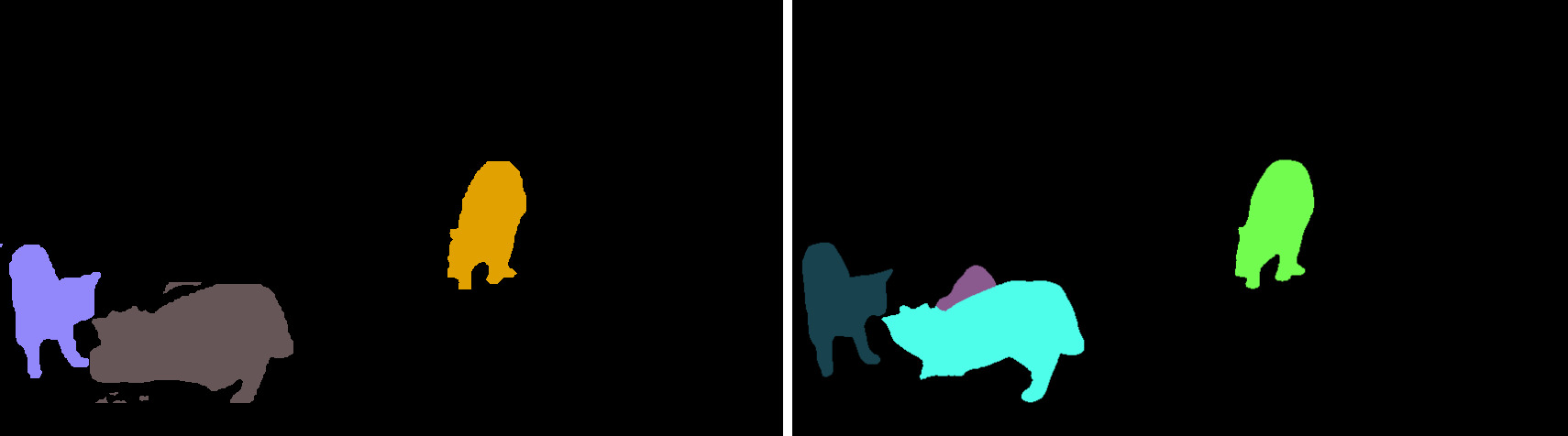}
\caption{\label{fig:5}Masks obtained through FastSAM and SAM2.}
\end{figure}

A comparison of the FastSAM and SAM2 segmentation models was performed using several key metrics. The test results are shown in Table \ref{tab:compare} and Figure \ref{fig:5}. The tests were performed on an RTX 2060 Super.

\begin{table}[h]
\centering
\caption{Comparison of segmentation methods}
\label{tab:compare}
\begin{tabular}{l|c|c|c|c|c}
\hline
Method & \begin{tabular}{@{}c@{}}Initializing \\model (ms)\end{tabular} & \begin{tabular}{@{}c@{}}Image \\Initialization \\(ms)\end{tabular} & \begin{tabular}{@{}c@{}}Mask \\prediction \\(ms)\end{tabular} & \begin{tabular}{@{}c@{}}VRAM\\ (MB)\end{tabular} & Prompts \\ 
\hline
FastSAM \cite{zhao2023fast} & 1357 & 379 & 15 & 607 & box, point \\
SAM2 \cite{ravi2024sam2} & 2722 & 660 & 50 & 1476 & box, point, both \\
\hline
\end{tabular}
\end{table}

The results of the comparative analysis demonstrate that SAM2, while being inferior to FastSAM in terms of total execution time (2722 ms vs. 1357 ms) and the amount of video memory used (1476 MB vs. 607 MB), outperforms it in terms of segmentation accuracy and provides more flexible functionality by supporting more types of prompts.

\section{Implementation}

Using the \textbf{SAM2} model provides flexible options for working with hints when performing image segmentation tasks. Cues can be represented in different formats such as points, rectangles or a combination of both. When using points, the user can specify parts of objects. This makes it possible to precisely define regions of interest and allows the user to specify regions of interest in different ways. The algorithm of Track Anything Annotate is divided into the following steps.

\textbf{Step 1: Initialize SAM2.} SAM2 \cite{ravi2024sam2} provides the ability to segment the region of interest using various hints. We use it to obtain an initial object mask. The user gets the mask of the object of interest in real time.

\begin{figure}[htbp]
\centering
\includegraphics[width=1\linewidth]{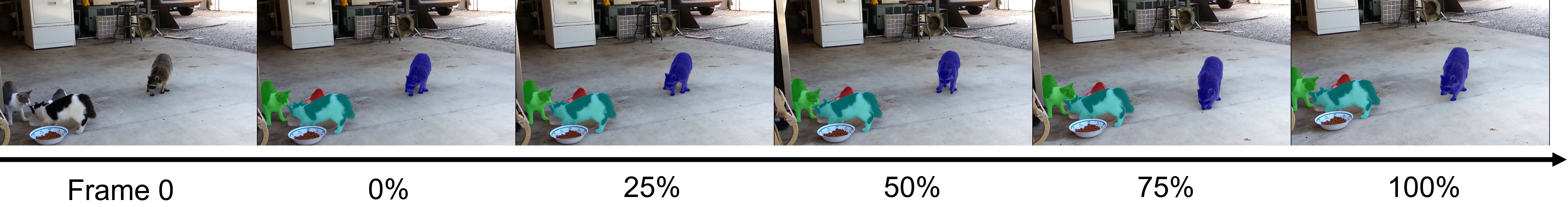}
\caption{\label{fig:6}Set of frames with superimposed masks obtained after XMem++ work.}
\end{figure}

\textbf{Step 2: Tracking with XMem++}. XMem++ \cite{bekuzarov2023xmem++} gets the mask and performs object segmentation on the following frames. In the output, we get masks of objects from each frame. Later we can overlay them on the original images for clarity. Figure \ref{fig:6} shows a set of frames with overlaid masks. They were obtained after XMem++ work.

\section{Tool prototype}

The prototype 'Track Anything Annotate' tool provides many possibilities for segmenting and annotating videos. Here we will demonstrate several scenarios of its use.

\begin{figure}[htbp]
\centering
\includegraphics[width=0.85\linewidth]{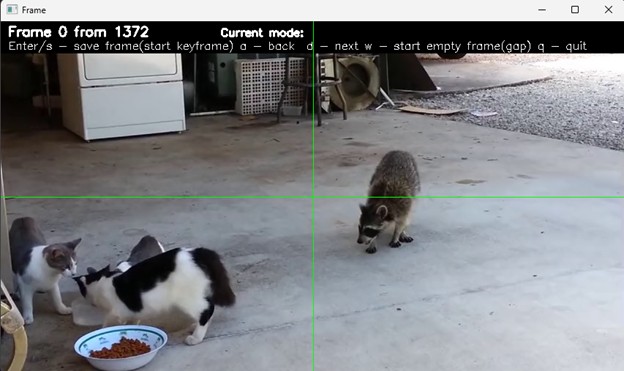}
\caption{\label{fig:7}Tool prototype interface.}
\end{figure}

\textbf{Creating a dataset based on segmentation of objects in the video}. The prototype has the ability to segment different sections of the video. You can do a full video segmentation or segment the video piece by piece. After that, a separate object tracking is performed and then a dataset is generated. The interface is shown in Figure \ref{fig:7}.

\begin{figure}[htbp]
\centering
\includegraphics[width=0.9\linewidth]{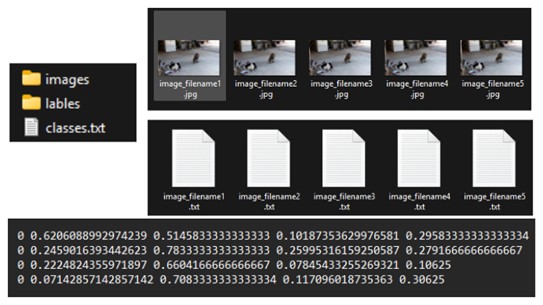}
\caption{\label{fig:8}Formed dataset.}
\end{figure}

\begin{figure}[htbp]
\centering
\includegraphics[width=1\linewidth]{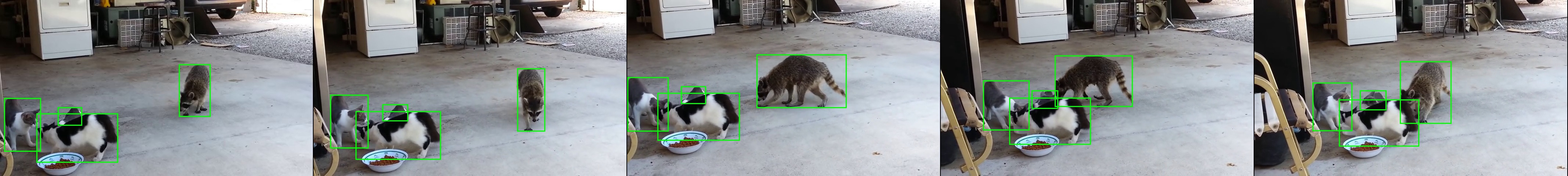}
\caption{\label{fig:9}An example of automatic markup of a generated dataset.}
\end{figure}

Dataset is formed in YOLO format. It includes text files with predefined classes and two folders: one with images and the other with text files, where the coordinates of objects are recorded. An example of the generated dataset is shown in Figure \ref{fig:8}. An example of automatic markup of the generated dataset is shown in Figure \ref{fig:9}.

\begin{figure}[htbp]
\centering
\includegraphics[width=0.95\linewidth]{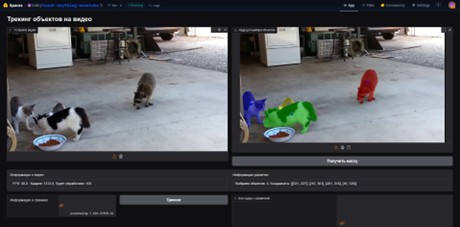}
\caption{\label{fig:10}Demo interface on HuggingFace.}
\end{figure}

\begin{figure}[htbp]
\centering
\includegraphics[width=0.95\linewidth]{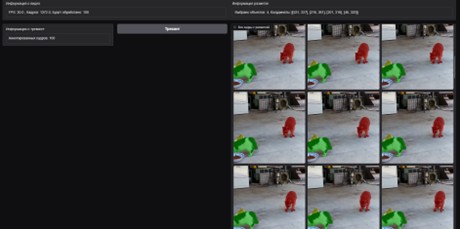}
\caption{\label{fig:11}Demo interface on HuggingFace.}
\end{figure}

\textbf{Web interface demo.} Here is a lightweight version of the application. It allows you to segment the first frame, view the mask, and run object tracking on the video. This creates a set of 100 frames with selected objects. A demo version of the application is available at Hugging Face\footnote{https://huggingface.co/spaces/lniki/track-anything-annotate}. The demo interface is shown in Figures \ref{fig:10} - \ref{fig:11}.

\section{Conclusion}

The developed tool solves the problem of labor-intensive data partitioning. For projects requiring long-term tracking and work with small objects, it is recommended to use the combination of SAM2 + Xmem++. This pair has proven to be the most effective despite high hardware requirements. The integration of SAM2 and Xmem++ not only accelerates annotation processes but also enhances accuracy in complex scenarios, such as overlapping objects or dynamic lighting conditions. Although hardware demands remain a limitation, the proposed approach significantly reduces manual intervention, making it a scalable solution for large-scale computer vision tasks.

\vspace{1cm}

The research was supported by the Ministry of Science and Higher Education of the Russian Federation within the framework of a state assignment (FEWZ-2024-0052).

\bibliographystyle{plain}
\bibliography{references}

\end{document}